\documentclass[conference]{IEEEtran}
\IEEEoverridecommandlockouts
\usepackage{cite}
\usepackage{amsmath,amssymb,amsfonts}
\usepackage{algorithmic}
\usepackage{graphicx}
\usepackage{textcomp}
\usepackage{xcolor}
\usepackage[numbers,sort&compress]{natbib}
\usepackage{balance}

\DeclareMathOperator*{\argmin}{arg\,min} % for argmin

\def\BibTeX{{\rm B\kern-.05em{\sc i\kern-.025em b}\kern-.08em
    T\kern-.1667em\lower.7ex\hbox{E}\kern-.125emX}}
    
\begin{document}

\title{EXPLORING GRAPH CLASSIFICATION TECHNIQUES
UNDER LOW DATA CONSTRAINTS: A COMPREHENSIVE STUDY\\
{\footnotesize \textsuperscript{}}}

\author{\IEEEauthorblockN{Kush Kothari\IEEEauthorrefmark{1},
Bhavya Mehta\IEEEauthorrefmark{2}, Reshmika Nambiar\IEEEauthorrefmark{3} and
Seema Shrawne\IEEEauthorrefmark{4}}
\IEEEauthorblockA{Department of Computer Engineering and Information Technology,
Veermata Jijabai Technology Institute\\
Mumbai, India.\\
Email: \IEEEauthorrefmark{1}kmkothari\_b19@ce.vjti.ac.in,
\IEEEauthorrefmark{2}bdmehta\_b19@ce.vjti.ac.in,
\IEEEauthorrefmark{3}rsnambiar\_b19@ce.vjti.ac.in,
\IEEEauthorrefmark{4}scshrawne@ce.vjti.ac.in}}
\maketitle

\begin{abstract}
This survey paper presents a brief overview of recent research on graph data augmentation and few-shot learning. It covers various techniques for graph data augmentation, including node and edge perturbation, graph coarsening, and graph generation, as well as the latest developments in few-shot learning, such as meta-learning and model-agnostic meta-learning. The paper explores these areas in depth and delves into further sub classifications. Rule based approaches and learning based approaches are surveyed under graph augmentation techniques. Few-Shot Learning on graphs is also studied in terms of metric-learning techniques and optimization-based techniques. In all, this paper provides an extensive array of techniques that can be employed in solving graph processing problems faced in low-data scenarios.
\end{abstract}

\begin{IEEEkeywords}
Graph Neural Networks, Graph Data Augmentation, Few Shot Learning.
\end{IEEEkeywords}

\section{Introduction}
Graphs are data structures that consist of a set of nodes or vertices that are interconnected using edges. Graphs are used in computer science and social sciences to model relationships and track the connections between entities. Since many real-life applications can be modeled using graphs, the available graph data is increasing exponentially. Today, graph data can be obtained in a wide array of applications like social network analysis, network route analysis, image processing and even fields like bioinformatics.

Hence, in recent times there has been a lot of development of efficient and accurate methods of graph data processing. Most recently, Graph Neural Networks (GNNs) \cite{gnn} are being used to perform convolution steps over graphs and extract useful information from graphs through node, edge and graph-level embeddings.

There are also certain open problems to consider when using GNNs to process graph data. The first is the overfitting of GNNs. When the training data consists of graphs with lots of nodes and edges, the high number of features in the data can cause the GNNs to easily overfit to a particular feature. This is detrimental to the testing performance of the Graph Neural Network. Second, the lack of labelled graph data in many scenarios poses another problem. In applications like molecular property prediction, it is usually not possible to obtain enough data due to the difficulty of manufacturing and testing new chemical compounds.

Thus there is a need to study graph processing methods that work well in low data scenarios while also preventing overfitting. Our contributions in this paper are:
\begin{enumerate}
\item A thorough study of Graph Data Augmentation (GDA) techniques.
\item A thorough study of Few Shot Learning (FSL) techniques for graph classification.
\end{enumerate}

\section{Background for Graph Classification}
 
\subsection{Introduction to Graph Neural Networks}
GNNs encode node features into low-dimensional space and learn representation vectors for the entire graph as well as for each individual node. A GNN architecture's main goal is to learn an embedding with neighbourhood information.

In general, an aggregation function (AGGREGATE) and an update function (UPDATE) can be used tfor establishing a general model for message passing GNNs. When updating each node's embedding in a layer, UPDATE integrates its own previous embedding with the aggregated neighbour embeddings. AGGREGATE combines the representations from the preceding layer for every single node from all of its neighbors. Specifically,

\begin{center}
  $
R^{i}_{\mathcal{N}(v)} = \text{AGGREGRATE}\left(\left\{ R_{u}^{i-1} | u \in \mathcal{N}(v) \right\}\right)
$
$
R^{i}_{v} = \text{UPDATE}\left(R_{v}^{i-1}, h_{\mathcal{N}(v)}^{i}\right)
$  
\end{center}

where at the \textit{i}-th GNN layer, \(R^{i}_{v}\) denotes the representation vector of node \textit{v}  and \(\mathcal{N}(v)\)specifies an organised set of a node \textit{v}'s neighbours.

\subsection{Graph Classification}
Graph classification mainly focuses on predicting an attribute for each graph in the collection by employing a supervised learning approach. Consider a graph represented by \(G(V,E)\), with graph label \textit{Y}, where \textit{V} denotes the set of vertices and \textit{E} is the set of edges, graph classification algorithms work on a function \(F: G \rightarrow Y\), where lowering the difference between the graph's true label and the forecasted labels is the primary objective of the classification method.

\section{Augmentation of Graph Data}
\begin{figure}[t]
    \centering
    \includegraphics[width=3.5in,height=7cm]{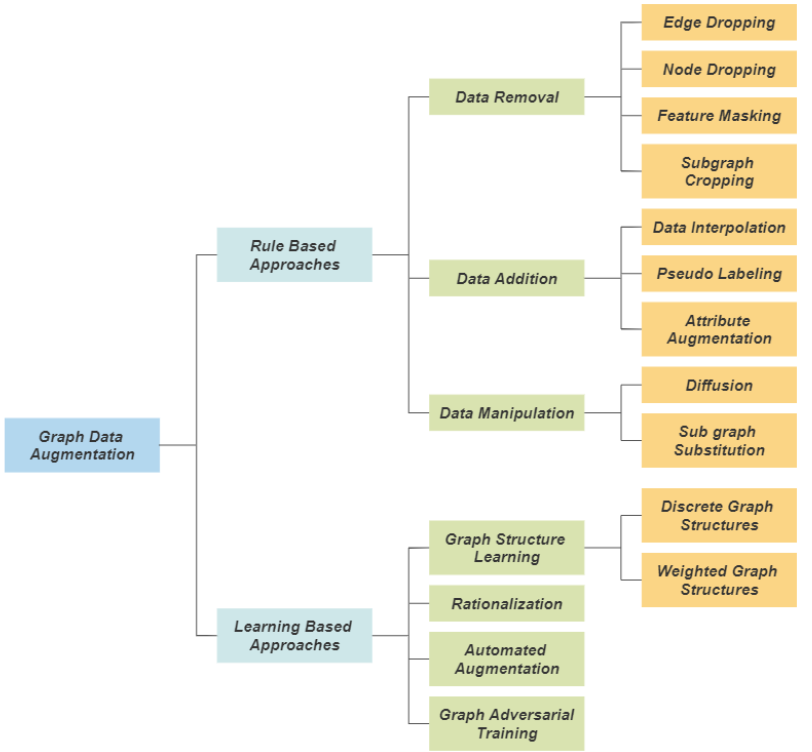}
    \caption{A tree level summary of existing Graph Data Augmentation Methods.}
    \label{fig:gda_overview}
\end{figure}

Augmentation methods  for graph data carry out denoising, imputation and strengthening of graph structure which help them better to match the goals or model processes of a target learning activity. Existing methodologies for GDA are mainly classified into two subparts namely, Rule Based approaches and Learning Based approaches ( Refer Fig.\ref{fig:gda_overview} )

% ===========================
% ===========================
\subsection{Rule Based Approaches}
Rule based approaches are straightforward and designed to preserve the essential characteristics of the original graph while introducing new variations. They are extensively used are mainly subdivided into three categories: \vspace{1.5pt}

\subsubsection{Data Removal}
\begin{itemize}
    \item Edge Dropping:
      \citeauthor{DropEdge}'s initial proposal of DropEdge \cite{DropEdge}, is similar to Dropout in Dense Neural Networks, arbitrarily drops a predetermined number of edges in every training epoch, addressing the over-smoothing problem of GNNs. Other significant methods like \citeauthor{thakoor}’s Bootstraping and \citeauthor{pmlr-v198-zhao22a}’s AutoGDA introduce different graph versions for every epoch in the training phase, improving the model’s generalization on deeper GNNs.

    \item Node Dropping: 
    DropNode\cite{dropNode} and NodeDropping\cite{Nodedropping} both seek to arbitrarily remove a portion of the nodes from the supplied graph, presuming that the missing nodes shouldn't have any impact on the network's overall semantics. \citeauthor{dropNode} in \cite{dropNode} employs a consistency loss on the projected logits of several enhanced versions of the graphs for augmentation tasks while \citeauthor{Nodedropping} in \cite{Nodedropping} concentrates on contrastive self-supervised graph representational learning.
    \item Feature Masking:
    A certain percentage of features are randomly removed from the source graph to create a new augmented graph. \citeauthor{dropmessage}'s DropMessage \cite{dropmessage} uses message passing GNNs to mask the features. FLAG by \citeauthor{flag} \cite{flag}, augments node features by iteratively adding gradient-based perturbations.
    \item Subgraph Cropping:
    We extract a sub-graph from a graph and perform cropping, swapping and modification operation on that subgraph.\citeauthor{imixup}'s ifMixup \cite{imixup} algorithm applies mixup by randomly assigning indices to nodes followed by index based node matching. Graph Transplant by \citeauthor{graphtransplant} \cite{graphtransplant}, first chooses a sub-graph from each source graph based on its salient nodes and is subsequently transplanted into the destination graph. Similar to this, \citeauthor{gmixup} \cite{gmixup} introduced G-Mixup, which first creates a class-level graph using graphons before combining those graphons to produce new data.
\end{itemize}

\subsubsection{Data Addition}
\begin{itemize}
    \item Data Interpolation: 
    By interpolating two annotated training samples, \citeauthor{mixup}’s Mixup generates virtual training instances adding prior information of the feature vectors and labels as given by:
    
    $$
    \Tilde{x} = \alpha x_{i} + (1 - \alpha)x_{j}
    $$
    $$
    \Tilde{y} = \alpha y_{i} + (1 - \alpha)y_{j}
    $$
    
    where \((x_{i},y_{i})\) and \((x_{j},y_{j})\) are two labelled training instances that were randomly chosen, and \(\alpha \in [0, 1]\).

    \citeauthor{manifold_mixup}'s Manifold Mixup uses latent intermediate representations rather than the two training samples' raw features. In contrast, Graph Mixup in \cite{gmixup} merges the raw attributes of each layer by passing the data to a two layer GNN and then combining them with vectors of each hidden layer. 
    % However Graph Mixup in \cite{gmixup} blends the unprocessed attributes of every layer, feeds them into the two-branch GNN layer, and blends the representations of each layer that are hidden.
    
    % Graph Mixup also works for the task of graph classification.  ifMixup \cite{imixup} directly applies Mixup on the graph data instead of the latent space for graph-level tasks. Graph Transplant\cite{graphtransplant} uses substructures as mixing units to preserve the local structural information. 
    % One more such algorithm by SMOTE interpolates examples within the minority classes which is especially effective when dealing with imbalanced data.

    \item Pseudo Labeling:
    Human labeling of graph data is expensivea and thus, pseudo-labeling is implemented as a semi-supervised approach in order to learn underlying contexts. According to the theory that adjacent nodes are more likely to share the same label, Label propagation \cite{sudolabelling1, sudolabelling2} iteratively distributes node labels down the edges. The introduced labels on the previously unlabeled nodes can then be used to train the GNN model.

    % Label propagation \cite{sudolabelling1, sudolabelling2} iteratively spreads node labels down the edges on the premise that related nodes are more likely to have the same label. The GNN model can then be trained with further labeled data using the propagated labels on the formerly unlabeled nodes.

    \item Attribute Augmentation:
    % Together with upgrading the graph topology, various works suggested creating new node properties to improve the graph data. 
    The Jumping Knowledge Network by \citeauthor{jkn} can be used to perform attribute augmentation for graph data which learns node embeddings that capture information from multiple layers of the graph. Another example of this type of work is LA-GNN by \citeauthor{lagnn}, which develops node attributes based on the conditional assignment of local patterns to enhance node localization.
    
    % Other such work is LA-GNN by \citeauthor{lagnn} which creates node features based on the conditional distribution of the local structures and neighbor features improving overall node locality.
    
\end{itemize}

\subsubsection{Data Manipulation}
\begin{itemize}
    \item Diffusion: 
    The basic idea is to propagate information from neighboring nodes to each node in the graph, based on the structure of the graph and the attributes of the nodes.
    \citeauthor{laplccian} in \cite{laplccian} constructs a Laplacian matrix from the graph adjacency matrix and then uses diffusion to smooth the node attributes over the graph. More recently, methods such as Heat Kernel Signature \cite{hks} and DeepWalk \cite{deepwalk} have been proposed to carry out the data diffusion task across edges and nodes.
    
    % Specifically, generalized graph diffusion is formulated as:
    % $$
    % \Tilde{A} = \sum^{\infty}_{k=0} \theta_{k} T^{k}
    % $$
    % where \(\theta_{k}\) denotes the global-local coefficient and \(T \in 	\mathbb{R}^{N \times N}\) represents the transition matrix derived from the adjacency matrix A.
\end{itemize}
    
\subsection{Learning based approaches}
% Rule based  approaches could sometimes be suboptimal since the augmentations do not take advantage of the rich information from downstream tasks. To address this concern, learned GDA approaches are proposed to learn augmentation strategies in a data-driven manner. 

Learning based GDA methods take advantage of rich information from downstream tasks to propose augmentation strategies in a data-driven manner.
\vspace{1.5pt}

\subsubsection{Graph Structure Learning (GSL)}
% In order to address situations when the graph-structured data is noisy or missing, graph structure learning has recently been revived in the context of GNNs.

\begin{figure}[t]
    \centering
    \includegraphics[width=3.48in]{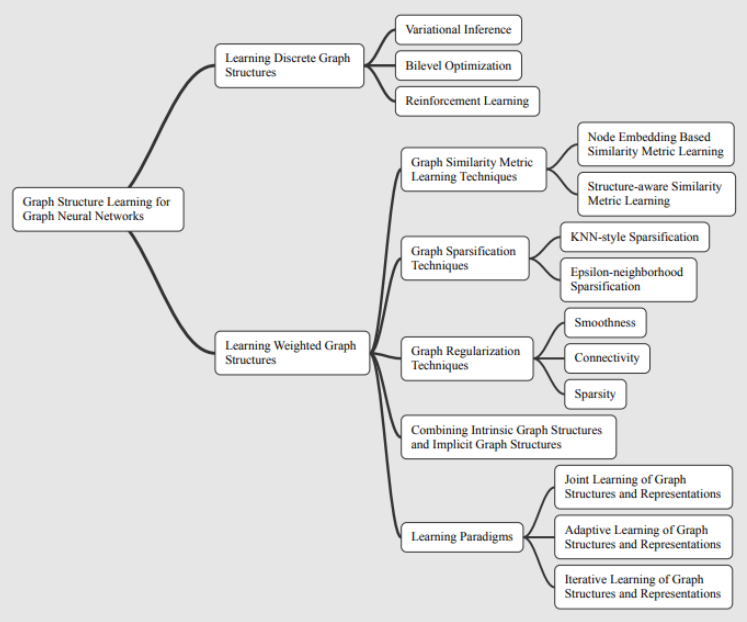}
    \caption{Graph Structure Learning methodologies.}
    \label{fig:gsl}
\end{figure}

Current endeavors in this field of study have mostly concentrated on automating the collaborative learning of graph structures without the aid of humans or subject-matter expertise. GSL methods are used when data is noisy or missing and based on the adjacency matrix they're trying to comprehend, as shown in Fig.\ref{fig:gsl}, they can be split into two groups.
% based on the adjacency matrix which they learn as shown in Fig.\ref{fig:gsl}.

% ===========================
% ===========================

\begin{itemize}
    \item Continuous structure learning:
    Learning a weighted graph structure helps gather rich information about edges. As compared to a binary matrix, a weighted continuous adjacency matrix is much easier to optimise because the latter can be done with Stochastic Gradient Descent or even convex optimisation techniques. Building upon this, the concept is further divided into five subparts as shown in Fig.\ref{fig:gsl}.

    \item Discrete structure learning: 
     A probabilistic adjacency matrix is used to derive a discrete graph structure. It mainly includes methods like bilevel optimisation \cite{bilevel}, Reinforcement Learning \cite{Kazi2023}, and variational inference \cite{pmlr-v80-chen18k, varinf} for the optimization task.
\end{itemize}

\subsubsection{Rationalization}
A subpart of the input attributes that represents, directs, and supports model prediction in the best possible manner is referred to as a rationale. They are typically naturally learnt subgraphs that serve as augmented graph data for the graph models and are representational, informative, or explanatory either employed alone or in conjunction with the original graph. CIGA was suggested by \citeauthor{ciga} in \cite{ciga} models the production of graphs addressing the out-of-distribution issues with structural causal models and enhances the interactions between invariant and spurious characteristics. A novel augmentation technique called Environment Replacement is by \citeauthor{GREA} in \cite{GREA}, which works by separating reasons from context.

\subsubsection{Automated Augmentation} 
Some recent research uses reinforcement learning techniques as a solution because automated GDA targets are frequently difficult to optimize. AutoGRL by \citeauthor{autogrl}, automatically learns the ideal blend of GDA actions, GNN architecture, and hyperparameters over the training procedure. AutoGDA by \citeauthor{autogda} in \cite{autogda} uses an RL-agent to develop localised augmentation strategies for node classification tasks and to generalise the learning. Another set of automated augmentation studies by Kose and Shen \cite{automated1}, Wang \cite{automated2} and others focus on graph contrastive learning.

\subsubsection{Graph Adversarial Training}
Machine learning models trained on graph data can become more robust and general by using a technique called "graph adversarial training". The main idea is to include erroneous predictions made by the machine learning model into the training data. The model can become more resilient to changes in the input data by learning from both normal and adversarial cases during training. 

% Unlike graph structure learning, graph adversarial training does not seek to find an optimal graph structure. To augment the adjacency matrix, \citeauthor{gat1} proposed to randomly drop edges during adversarial training without any optimization on the graph data. 

Contrary to learning the ideal graph structure, graph adversarial training does not look for one. \citeauthor{gat1} in \cite{gat1} proposed a method which randomly deletes edges during adversarial training without any optimisation on the graph data in order to augment the adjacency matrix. An adversarial training technique with dynamic regularisation was put forth by \citeauthor{gat2} in \cite{gat2}, with the goal of restoring the uniformity of the graph and limiting the disparity between the anticipated outcomes of the linked nodes and the target node.
% reconstructing graph smoothness and constraining the divergence between the target node's prediction and the predictions of its linked nodes. Furthermore, even FLAG \cite{flag} uses adversarial training to progressively supplement the node characteristics with gradient-based adversarial perturbations.
Even FLAG\cite{flag} uses adversarial conditioning to progressively incorporate adversarial gradient based permutations to corresponding node properties.

\subsection{M-Evolve}

\citeauthor{mevolve} in \cite{mevolve} proposed M-Evolve, a framework which optimizes pre-trained graph classifiers using an evolutionary method by combining the augmentation of graphs, filtering out bad graphs and retraining the model.

\textbf{Graph augmentation.} They generated graphs with modified edges by adding and removing $\lceil \beta \cdot m \rceil$ edges, where $m$ represents the total number of edges in the original graph and $\beta$ denotes the budget of edge modification. To select edges for addition to/removal from the graph, they defined a candidate set for each, $E_{add}^c$ and $E_{del}^c$ respectively.

The construction of these candidate sets was done via two methods: random mapping (for a simple baseline) and motif similarity mapping. In motif similarity mapping, repeating patterns in the graph (called graph motifs) are modified in some way. M-Evolve uses open-triads as the motif. An open-triad $\wedge_{ij}^a$ represents a graph structure with a path of length 2 connecting vertex $v_{i}$ to $v_{j}$ when passing through $v_{a}$ such that there is no direct edge between $v_{i}$ and $v_{j}$ (See Figure 3). The candidate sets identify such open-triad motifs and modify them such that the missing edge between vertices $v_{i}$ and $v_{j}$ is added to $E_{add}^c$, and the existing two edges of the open-triad are added to $E_{del}^c$.

% ===========================
\begin{figure}[t]
    \centering
    \includegraphics[width=1.48in]{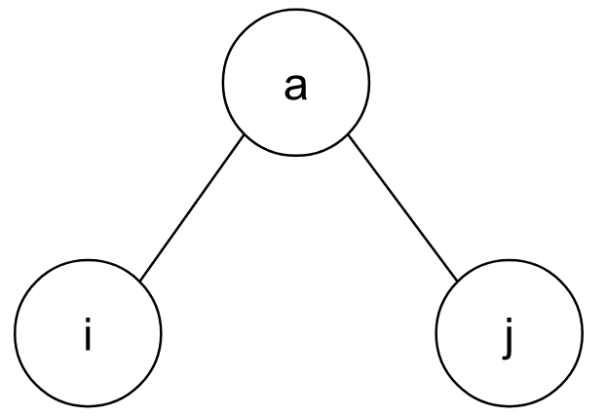}
    \caption{An open triad, $\wedge_{ij}^a$. M-Evolve modifies such triads in a graph.}
    \label{fig:mevolve}
\end{figure} 
% ===========================

Edges are selected from the candidate sets via weighted random sampling. The weights for this are calculated by finding the similarity between two vertices. The closer they are, the more likely it is to add an edge between them, and vice versa. The similarity has been found using Resource Allocation index ($s_{ij}$).

\[s_{ij} = \sum_{k\in \eta( \,i) \,\cap\eta( \,j) \,}\frac{1}{d_{k}}, \quad S = \{s_{ij}|\forall ( \,v_{i},v_{j}) \, \in E_{add}^c \cup E_{del}^c\}\]
where $\eta( \,i) \,$ represents the neighbors of $v_{i}$ (within one-hop), and $d_{k}$ denotes vertex $k$'s degree. Weights are then calculated.

\[w_{ij}^{add} = \frac{s_{ij}}{\sum_{s \in S}}, \quad w_{ij}^{del} = 1 - \frac{s_{ij}}{\sum_{s \in S}}\]

\textbf{Data filtration.} M-Evolve filters out augmented graphs with low label reliability. First, the classifier $C$ is pretrained on the initial training data ($D_{train}$) and validation data ($D_{val}$). For a specified number of evolution iterations $T$, augmentations are performed. A prediction vector $p_{i}$ is found for each graph $G_{i}$ in $D_{val}$, which is the probability of a class being correct for $G_{i}$. A confusion matrix for probability $Q$ is calculated where each entry has the mean probability ($q_{ij}$) of a graph being classified by $C$ as class $j$ instead of the correct class $i$. $q_{k}$ is the average probability of a graph belonging to class $k$.

Label reliability is calculated for each graph in $D_{val}$. For a graph $G_{i}$ in $D_{val}$ with $y_{i}$ as the correct label, the label reliability ($r_{i}$) of this example is calculated.

\[r_{i} = p_{i}^\top q_{y_{i}}\]

To filter out graphs from the generated pool of graphs $D_{pool}$, a threshold $\theta$ is found such that only graphs with $r_{i}$ above $\theta$ will be added to the new training set $D_{train}^{new}$.

\[\theta = \argmin_\theta\sum_{( \,G_{i},y_{i}) \, \in D_{val}} \Gamma[ \,( \,\theta - r_{i}) \,\cdot c( \,G_{i},y_{i}) \,] \,\]

Here, if $x > 0$, $\Gamma( \,x) \, = 1$ and it is $0$ otherwise. If $C( \,G_{i}) \, = y_{i}$, then the value of $c( \,G_{i},y_{i}) \, = 1$ and it is $-1$ otherwise.

The \textbf{evolutionary classifier} $C$ is then retrained using data added at each evolutionary iteration.

M-Evolve using the motif similarity-based mapping performed better than several graph classification methods. They observed an increase in the dataset scale, smoother decision boundaries, and less fragmented decision regions.

\section{Few Shot Learning based methods}
Even without directly augmenting the domain-specific and/or task-specific data, previous research has been conducted that applies various specialized techniques on low-data scenarios. A lot of this research has been focused on general data types, but research has also been conducted in graph data specific domains. Such techniques often deal with low-data tasks and are tuned to generalize well on previously unseen data and classes. These techniques are often referred to as few-shot learning algorithms. In the sections that follow, special attention has been paid to few-shot learning algorithms that are general enough to work well with graph data or are specifically designed for graph data.

It is rather important to note that while these few-shot learning algorithms are being studied in isolation with data augmentation techniques, in practice a combination of both would yield much better results. However, the question of which augmentation methods to match with which few-shot algorithms is a multi-faceted one, and it often boils down to the domain of the data and the quality and type of data available for the tasks.

A thorough dive into FSL literature that could be adapted for the graph classification task follows in the sections below. We first state the few-shot graph classification problem objectively and then provide a rough nomenclature of the available graph classification techniques as they have been described in general FSL reviews like \cite{DBLP:journals/corr/abs-1904-04232} and \cite{10.1145/3386252}.

This classification divides the techniques into three main categories:
\begin{enumerate}
    \item \textbf{Model based techniques:} Models that train well on the few-shot data are often augmented with memory units so that prior domain-based information can be stored. This information will be helpful when training on a specific task which does not require a large amount of data points.
    
    \item \textbf{Metric-Learning based techniques:} These techniques tackle a few-shot graph classification problem by “learning to compare” the inherent structure of graphs. Domain-specific graphs often have inherent patterns expressed in their structure and their classes. 
    
    % These techniques often train the parameters to identify and differentiate between such patterns.
    \item \textbf{Optimization based techniques:} These techniques provide a good initialization parameters for a model. Once we optimize the set of initial parameters in the model, it becomes much easier to fine-tune the model to a FSL task that has similar but limited data. This parameter optimization step is often done in a “meta-learning” stage that comes before the main training stage. Literature in these methods also focuses on methods specific to optimizing on nodes of a graph.
\end{enumerate}

\subsection{Metric-Learning based methods}
Metric-learning based methods exploit the fact that graphs that belong to a specific class often have similar structural patterns. Thus, if we are able to extract such structural features (or metrics) from graphs, it would be easier to compare two graphs that lie in the same or different classes. For this to occur, we need efficient ways to embed graphs in a feature space and compare two graphs to check if they have the same class label.

\subsubsection{Graph Kernel Methods}
Graph Kernel Methods aim to capture important features of the graph by quantifying its various properties. Most commonly, these Graph Kernels encode pairs of graphs at the same time and help identify how similar or different they are in overall structure. They provide mathematical formulae to capture an inner-product between graphs that quantifies this similarity. Graph Kernels were one of the first tools to be used to perform Machine Learning on graph data. Kernel-based learning algorithms like Support Vector Machines can now easily operate on graphs.

Work by \citeauthor{articlehaussler} in \cite{articlehaussler} is one of the earliest known instances in literature to have applied kernels on discrete structures like graphs. A fundamental type of graph kernel is the random walk kernel, that performs random walks on the direct product of the pair of graphs. This product graph is as defined in the equation below

Another notable type of graph kernel are the shortest-path graph kernels introduced in \cite{1565664} by \citeauthor{1565664}. They improve upon random walk kernels by providing pre-computed shortest paths via the Floyd-Warshall algorithm. However, these are not ideal in cases where longest paths and average paths are more adequate. Thus, designing graph kernels tailored for specific tasks was still a necessity.

\citeauthor{JMLR:v12:shervashidze11a} probed further and introduced a family of efficient graph kernels that provided quick feature extraction for large graphs. The Weisfeiler-Lehman graph isomorphism test served as the foundation for this feature extraction method. These kernels are generally accepted to be the current state-of-the-art for graph classification.

\subsubsection{SuperClass}
Since graph kernels have limited metric learning abilities and often require extensive computation to compute, they cannot be relied upon in every scenario. Instead, to determine patterns between graphs of different classes, \citeauthor{DBLP:journals/corr/abs-2002-12815} in \cite{DBLP:journals/corr/abs-2002-12815}, used spectral graph theory to determine the patterns of connectivity in the discrete structures. Spectral analytics could help identify patterns within a single graph, however to identify patterns between different classes, \citeauthor{DBLP:journals/corr/abs-2002-12815} performed graph clustering in the spectral dimension with the Lp Wasserstein distance as the distance metric. A prototype graph of each class is first generated using the mean spectral measure of each data point belonging to the graph. After that, the individual class prototypes are then clustered together using K-means++ \cite{10.5555/1283383.1283494}. Each of these clusters is assigned a super-class.

\subsubsection{CuCO}
\citeauthor{ijcai2021p317} in \cite{ijcai2021p317} proposed a new framework called CuCo (using curriculum contrastive learning) for learning graph representations, with a special focus on negative sampling. It was an unsupervised/self-supervised learning approach for applications with limited labeled data.

The authors used four basic data augmentation techniques: dropping of nodes, perturbing edges, masking attributes, and sampling certain subgraphs. They strategically selected the node dropping and subgraph sampling techniques for molecular data, and used all four techniques for social network data. The model uses GNNs for the Graph Encoder which learns graph representations.

Contrastive learning, a self-supervised algorithm, is often used for representation learning. The process involves performing augmentations on a data instance, pairing them up, and labeling them as positive pairs, while pairing up two dissimilar data instances to get negative pairs. It then forces the embeddings of the instances of a positive pair to be closer to each other, and the embeddings of the instances of negative pairs to be farther apart.

CuCo implements this by augmenting a graph \(\mathcal{G}\) to get two generated graphs/views as a positive pair, while any other graph in the training set is paired with \(\mathcal{G}\) to get the negative pair. Memory bank \(\mathcal{Q}\) of size \(\mathcal{K}\) represents all the negative samples in the training set, i.e., all the graphs except \(\mathcal{G}\). A similarity metric function (dot product in this case) sim(·, ·) is employed on the pairs of embeddings. Based on this, the noise-contrastive estimation loss function is used as follows:

where \(z_{i} \) is the embedding/representation of graph \(\mathcal{G}_{i}\), and \{\(z_{i},z_{j}\)\} and \{\(z_{i},z_{k}\)\} are the positive pairs and negative pairs respectively. $\tau$ denotes the temperature parameter. This loss ensures that the positive pairs score higher in terms of similarity than the negative pairs.

Curriculum learning is a training strategy where a machine learning model is trained from samples which are ordered by their level of difficulty. Easier samples are learned first and harder ones are sampled in the later stages of training. This is shown to improve the generalization capability of such models.

\citeauthor{ijcai2021p317} realised the impact of negative sampling and thus implemented curriculum learning to sample negative pairs strategically. Similarity was used as the scoring function to measure the difficulty of negative samples. The more similar two graphs are, the harder it will be to differentiate between them and classify the sample as a negative pair. Cosine similarity and dot product similarity were employed for this.

To confirm the benefits of using curriculum learning, the model was also tested using random ordering of samples, as well as an anti-curriculum order which feeds negative samples in the descending order of difficulty. They found that the curriculum order (ascending order of difficulty) performed best.

In addition to these methods, they added an early stop mechanism with patience $\pi$ which will reduce whenever loss does not decrease. Due to ordered learning and the early stop mechanism, CuCo was found to learn faster.

The model was tested under the unsupervised representation learning setting and the transfer learning setting. The unsupervised learning evaluation was done on graph classification tasks where the embeddings learnt by the model were used to train an SVM classifier. The graph encoder used was a three-layer GIN. The datasets used for unsupervised learning were obtained from the TUDataset. CuCo outperformed the baselines on 6 out of 7 of the chosen datasets. CuCo was also found to be effective for transfer learning by achieving the best results on six of eight molecular property prediction datasets of the OGB benchmark.

\subsection{Optimization-based methods}
Another class of techniques are ones that provide good initialization model parameters. These initial parameters are trained and optimized in a step before the main on-task training of the model, in a step called the meta-training stage. Meta-training is often carried out on tasks that are similar to the low-data task. This allows us to introduce an inductive bias on the set of parameters. This ensures that our parameters are initialized in such a way that they can converge quickly to the optimal set of parameters when the model is trained on low-data tasks. Another concern is to also prevent the overfitting of the model parameters during the fine-tuning step.

In the following sections we mostly look at \cite{10.5555/3305381.3305498} and its derivative work by \citeauthor{DBLP:journals/corr/abs-2003-08246} which adapts techniques used in \cite{10.5555/3305381.3305498} specially to graphs.

\subsubsection{Model Agnostic Meta Learner}
The overarching goal of Few-Shot Learning is to provide Machine Learning algorithms the ability to generalize well to a small amount of data just like humans without overfitting to small datasets. \citeauthor{10.5555/3305381.3305498} in \cite{10.5555/3305381.3305498} presents a very general task structure and model agnostic algorithm that is able to provide state-of-the-art results on the initialization of parameters during the meta-learning stage. The method is general and independent of the model architecture or the problem type (classification, regression, or reinforcement learning). The only restriction on the model is that it must be optimizable by some gradient-based method. Hence, this method can be readily used on models consisting of fully-connected neural networks, convolutional neural networks, and in our use case, even graph neural networks.

In this technique, instead of directly training the parameters, we alter the initial parameters in such a way that maximum performance will be achieved in minimum steps of gradient descent on the novel task. In other words, we train the initial parameters such that they will work well when fine tuned on the data of the task during the training stage. This gives the desired result that ensures that a minimum number of steps of gradient descent are required at the training stage (also called fine-tuning stage in some derivative literature).

MAML achieves these goals, by using the model parameters to build an internal representation of the tasks and learn features of the data that is often shared across these tasks. Since task-agnostic features are already being extracted at the meta-learning stage, the model becomes more sensitive to task-specific features during the training stage. This is described as rapid task adaptation in the paper.

This shifts the paradigm of pretraining away from finding a set of initial parameters that perform well on all tasks, towards optimizing for parameters that converge on task-specific optimality as quickly as possible.

\subsubsection{AS-MAML}
% ===========================
\begin{figure}[t]
    \centering
    \includegraphics[width=3.48in]{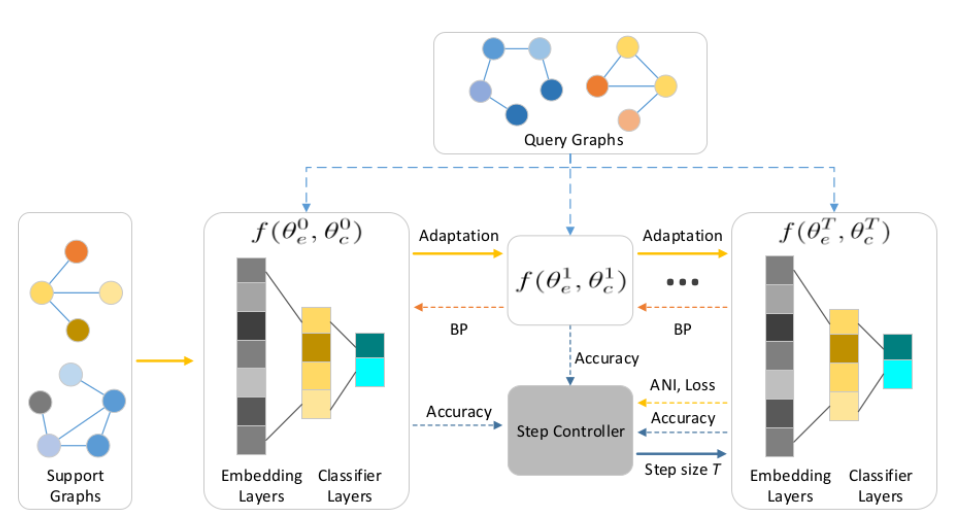}
    \caption{Generic AS-MAML Architecture as given in \cite{DBLP:journals/corr/abs-2003-08246}}
    \label{fig:asmamal-arch}
\end{figure} 
% ===========================
An important issue with approach taken by \citeauthor{10.5555/3305381.3305498} is that the learning rate hyperparameters alpha and beta (for the parameter optimization and meta optimization) are very difficult to determine and often have major effects on the performance of MAML. \citeauthor{DBLP:journals/corr/abs-2003-08246} in \cite{DBLP:journals/corr/abs-2003-08246} takes a more automated approach to learning the step-size during the meta-learning phase.

The approach proposed by \citeauthor{DBLP:journals/corr/abs-2003-08246}(commonly referred to as AS-MAML) consists of two parts, the Adaptive Step Meta-Learner and a few-shot Graph classifier. The adaptive step meta learner is responsible for adjusting the GNN’s step size and adjusting how much information is learnt from each gradient step based on the loss function gradient and the current iteration number. The overall architecture is explained in Fig.\ref{fig:asmamal-arch}. Reinforcement learning was applied to provide the optimal adaptation step for meta-learner using the quality of the graph embedding.
The quality of the graph embedding was determined using the Average Node Embedding (ANI) value, where an increase in ANI value signifies an increase in the quality of graph embedding that was created.

\section{Conclusion}
In conclusion, this research paper presents a detailed survey of graph classification techniques in low data scenarios. The review focusses on two major solutions to low data contraints: Graph Data Augmentation techniques and Few-Shot Learning Graph Algorithms. We researched both rule-based and learning-based GDA approaches and discussed some specific algorithms like M-Evolve. Lastly, we covered few-shot learning techniques specially designed for graphs like metric-Learning based methods (eg Graph Kernel Methods and CuCO) and optimization-based methods (like MAML and its derivative AS-MAML).

{
\scriptsize
\bibliographystyle{IEEEtranN}
\bibliography{bibliography}
}
\balance

\end{document}